\documentclass{article} 
\usepackage{iclr2024_conference,times}
\usepackage[utf8]{inputenc} 
\usepackage[T1]{fontenc}    
\usepackage{hyperref}       
\usepackage{url}            
\usepackage{booktabs}       
\usepackage{amsfonts}       
\usepackage{nicefrac}       
\usepackage{microtype}      
\usepackage{xcolor}         
\usepackage{graphicx}       
\usepackage{subcaption}     
\usepackage{textcomp}
\usepackage{amsmath}
\usepackage{amssymb}
\usepackage{multirow}
\usepackage{siunitx}
\usepackage{tikz}
\usetikzlibrary{shapes.misc}
\tikzset{cross/.style={cross out, draw=black, minimum size=2*(#1-\pgflinewidth), inner sep=0pt, outer sep=0pt},
cross/.default={4.5pt}}
\usepackage{enumitem}
\usepackage{algorithm}
\usepackage[noend]{algpseudocode}

\usepackage[colorinlistoftodos]{todonotes}
\usetikzlibrary{shapes,arrows,positioning,fit,calc,backgrounds,shapes.geometric}

\definecolor{babyblue}{rgb}{0.06,0.58,0.97}
\definecolor{salmon}{rgb}{0.98,0.41,0.38}

\graphicspath{ {./figures/} }

\usepackage{amsmath,amsfonts,bm}

\def\eqref#1{equation~\ref{#1}}

\def\1{\bm{1}}

\DeclareMathAlphabet{\mathsfit}{\encodingdefault}{\sfdefault}{m}{sl}
\SetMathAlphabet{\mathsfit}{bold}{\encodingdefault}{\sfdefault}{bx}{n}

\newcommand{\figref}[1]{Figure~\ref{fig:#1}}

\newcommand{\figlabel}[1]{\label{fig:#1}} 
\newcommand{\tableref}[1]{Table~\ref{table:#1}}

\newcommand{\tablelabel}[1]{\label{table:#1}}
\newcommand{\algoref}[1]{Algorithm~\ref{algo:#1}}
\newcommand{\algolabel}[1]{\label{algo:#1}}
\newcommand{\sectionref}[1]{Section~\ref{section:#1}}
\newcommand{\sectionlabel}[1]{\label{section:#1}}
\newcommand{\appendixref}[1]{Supplemental~\ref{appendix:#1}}
\newcommand{\appendixlabel}[1]{\label{appendix:#1}}
\newcommand{\equationref}[1]{Equation~\ref{equation:#1}}
\newcommand{\equationlabel}[1]{\label{equation:#1}}

\newcommand{\ourpipeline}{SFvD}
\newcommand{\ourpipelinefull}{Scene Flow via Distillation}

\newcommand{\ourmethod}{ZeroFlow}
\newcommand{\xl}{XL}
\newcommand{\threex}{3X}
\newcommand{\fivex}{5X}
\newcommand{\twox}{2X}
\newcommand{\onex}{1X}
\newcommand{\ourmethodonex}{ZeroFlow \onex{}}
\newcommand{\ourmethodthreex}{ZeroFlow \threex{}}
\newcommand{\ourmethodfivex}{ZeroFlow \fivex{}}
\newcommand{\ourmethodtwox}{ZeroFlow \twox{}}
\newcommand{\ourmethodxlonex}{ZeroFlow \xl{} \onex{}}
\newcommand{\ourmethodxlthreex}{ZeroFlow \xl{} \threex{}}
\newcommand{\ourmethodfull}{Zero-Label Scalable Scene Flow}

\newcommand{\pointcloudt}{P_t}
\newcommand{\pointcloudtpone}{P_{t+1}}

\newcommand{\flow}{\hat{F}}
\newcommand{\flowttpone}{\hat{F}_{t,t+1}}

\newcommand{\flowgtttpone}{F^*_{t,t+1}}

\newcommand{\norm}[1]{\left\lVert #1 \right\rVert}

\newcommand{\poorparagraph}[1]{\textbf{{#1}.}}
\newcommand{\humanlabels}[1]{\underline{{#1}}}
\newcommand{\bgscale}[1]{\sigma(#1)}
\newcommand{\pointspeed}[1]{s(#1)}

\newcommand{\flowforward}{\flow^+}
\newcommand{\flowrev}{\flow^-}
\newcommand{\chamferdistancenameraw}{TruncatedChamfer}
\newcommand{\chamferdistancename}{\textup{\chamferdistancenameraw{}}}
\newcommand{\chamferdistance}[2]{\chamferdistancename(#1, #2)}

\newcommand{\bevmaintextfontsize}{\fontsize{5pt}{5pt}\selectfont}
\DeclareCaptionFont{bevmaintextfont}{\bevmaintextfontsize}

\newcommand{\resulttablefontsize}{\fontsize{6.3pt}{6.3pt}\selectfont}
\DeclareCaptionFont{resulttablefont}{\resulttablefontsize}

\makeatletter
\DeclareRobustCommand{\iscircle}{\mathord{\mathpalette\is@circle\relax}}
\newcommand\is@circle[2]{%
  \begingroup
  \sbox\z@{\raisebox{\depth}{$\m@th#1\bigcirc$}}%
  \sbox\tw@{$#1\square$}%
  \resizebox{!}{\ht\tw@}{\usebox{\z@}}%
  \endgroup
}
\makeatother

\newcommand{\tikzcircle}[2][black,fill=black]{\tikz[baseline=-0.5ex]\draw[#1,radius=#2] (0,0.03) circle ;}%

\newcommand{\tikzx}[2][black,fill=black]{\tikz[baseline=-0.5ex]\draw[#1] (0,0.03)  node[cross,rotate=0, line width=1.2pt] {}; }%

\usepackage{hyperref}
\usepackage{url}

\title{\textit{\ourmethod{}}: Scalable Scene Flow via Distillation}

\author{%
Kyle Vedder$^{1}$\thanks{Corresponding email: \texttt{kvedder@seas.upenn.edu}} \quad Neehar Peri$^{2}$ \quad Nathaniel Chodosh$^2$ \quad Ishan Khatri$^3$ \quad Eric Eaton$^1$ \\ \textbf{Dinesh Jayaraman}$^1$ \quad \textbf{Yang Liu$^4$} \quad \textbf{Deva Ramanan}$^2$ \quad \textbf{James Hays}$^5$\\
{\small$^1$University of Pennsylvania \quad $^2$Carnegie Mellon University \quad $^3$Motional} \\ {\small$^4$Lawrence Livermore National Laboratory \quad $^5$Georgia Tech}
}

\iclrfinalcopy 
\begin{document}

\maketitle

\begin{abstract}

Scene flow estimation is the task of describing the 3D motion field between temporally successive point clouds. State-of-the-art methods use strong priors and test-time optimization techniques, but require on the order of tens of seconds to process full-size point clouds, making them unusable as computer vision primitives for real-time applications such as open world object detection. Feedforward methods are considerably faster, running on the order of tens to hundreds of milliseconds for full-size point clouds, but require expensive human supervision. To address both limitations, we propose \emph{\ourpipelinefull{}}, a simple, scalable distillation framework that uses a label-free optimization method to produce pseudo-labels to supervise a feedforward model. Our instantiation of this framework, \emph{\ourmethod{}}, achieves \textbf{state-of-the-art} performance on the \emph{Argoverse~2 Self-Supervised Scene Flow Challenge} while using zero human labels by simply training on large-scale, diverse unlabeled data. At test-time, \ourmethod{} is over 1000$\times$ faster than label-free state-of-the-art optimization-based methods on full-size point clouds (34 FPS vs 0.028 FPS) and over 1000$\times$ cheaper to train on unlabeled data compared to the cost of human annotation (\$394 vs $\sim$\$750,000). To facilitate further research, we release our code, trained model weights, and high quality pseudo-labels for the Argoverse~2 and Waymo Open datasets at \url{https://vedder.io/zeroflow}.
\end{abstract}

\setcounter{footnote}{0} 

\section{Introduction}
Scene flow estimation is an important primitive for open-world object detection and tracking~\citep{objectdetectionmotion, Zhai2020FlowMOT3M, baur2021slim, huang2022accumulation, flowssl}. As an example, \citet{objectdetectionmotion} generates supervisory boxes for an open-world LiDAR detector via offline object extraction using high quality scene flow estimates from Neural Scene Flow Prior (NSFP) \citep{nsfp}. Although NSFP does not require human supervision, it takes tens of seconds to run on a single full-size point cloud pair. If NSFP were both high quality and real-time, its estimations could be directly used as a runtime primitive in the downstream detector instead of relegated to an offline pipeline. This runtime feature formulation is similar to \citet{Zhai2020FlowMOT3M}'s use of scene flow from FlowNet3D~\citep{flownet3d} as an input primitive for their multi-object tracking pipeline; although FlowNet3D is fast enough for online processing of subsampled point clouds, its supervised feedforward formulation requires significant in-domain human annotations. 


Broadly, these exemplar methods are representative of the strengths and weakness of their class of approach. Supervised feedforward methods use human annotations which are expensive to annotate\footnote{At $\sim$\$0.10 / cuboid / frame, the Argoverse~2~\citep{argoverse2} \emph{train} split cost $\sim$\$750,000 to label; \ourmethod{}'s pseudo-labels cost \$394 at current cloud compute prices. See \appendixref{labelvspseudolabelcosts} for details.}. To amortize these costs, human annotations are typically done on consecutive observations, severely limiting the structural diversity of the annotated scenes (e.g.\ a 15 second sequence from an Autonomous Vehicle typically only covers a single city block); due to costs and labeling difficulty, large-scale labels are also rarely even available outside of Autonomous Vehicle domains. Test-time optimization techniques circumvent the need for human labels by relying on hand-built priors, but they are too slow for online scene flow estimation\footnote{NSFP~\citep{nsfp} takes more than 26 seconds and Chodosh~\citep{chodosh2023} takes more than 35 seconds per point cloud pair on the Argoverse~2~\citep{argoverse2} train split. See \appendixref{labelvspseudolabelcosts} for details.}.

We propose \emph{\ourpipelinefull} (\ourpipeline{}), a simple, scalable distillation framework that uses a label-free optimization method to produce pseudo-labels to supervise a feedforward model. \ourpipeline{} generates a new class of scene flow estimation methods that combine the strengths of optimization-based and feedforward methods with the power of data scale and diversity to achieve fast run-time and superior accuracy without human supervision. We instantiate this pipeline into \emph{\ourmethodfull{}} (\ourmethod{}), a family of methods that, motivated by real-world applications, can process full-size point clouds while providing high quality scene flow estimates. 
We demonstrate the strength of \ourmethod{} on Argoverse~2 \citep{argoverse2} and Waymo Open \citep{waymoopen}, notably achieving \textbf{state-of-the-art} on the \emph{Argoverse~2 Self-Supervised Scene Flow Challenge} 
(\figref{tradeoff_curve}).

\begin{figure}[t]
  \centering
  \includegraphics[width=\linewidth]{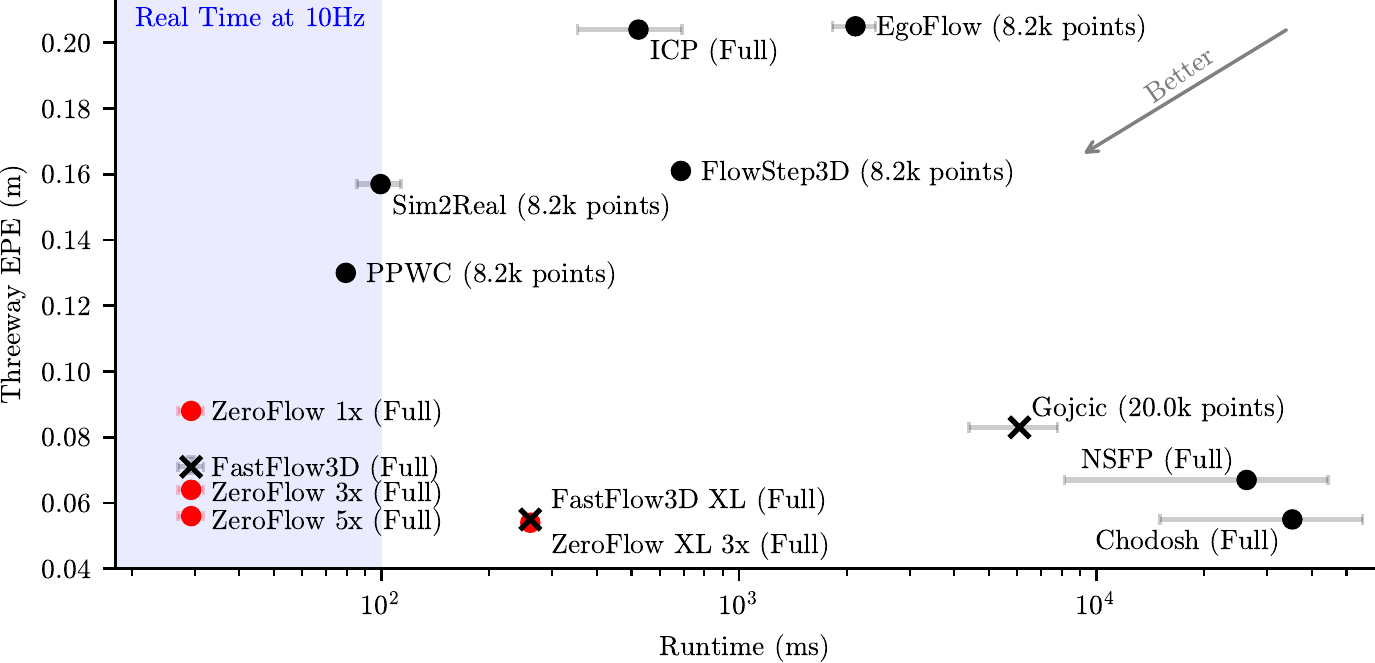}
  \caption{We plot the error and run-time of recent scene flow methods on the Argoverse~2 Sensor dataset~\citep{argoverse2}, along with the size of the point cloud prescribed in the method's evaluation protocol. Our method \textcolor{red}{\ourmethodthreex{}} (ZeroFlow trained on 3$\times$ pseudo-labeled data) outperforms its teacher (NSFP,~\cite{nsfp}) while running over 1000$\times$ faster, and \textcolor{red}{\ourmethodxlthreex{}} (ZeroFlow with a larger backbone trained on 3$\times$ pseudo-labeled data) achieves \textbf{state-of-the-art}. Methods that use \emph{any} human labels are plotted with \tikzx{3.5pt}, and zero-label methods are plotted with \tikzcircle{3.5pt}.  
} 
  \figlabel{tradeoff_curve}
\end{figure}


Our primary contributions include: 
\begin{itemize}[leftmargin=*]
  \item We introduce a simple yet effective distillation framework, \emph{\ourpipelinefull{}} (\ourpipeline{}), which uses a label-free optimization method to produce pseudo-labels to supervise a feedforward model, allowing us to surpass the performance of slow optimization-based approaches at the speed of feedforward methods.
  \item Using \ourpipeline{}, we present \emph{\ourmethodfull{}} (\ourmethod{}), a family of 
 methods that produce fast, \textbf{state-of-the-art} scene flow on full-size clouds, with methods running over 1000$\times$ faster than state-of-the-art optimization methods (29.33 ms for \ourmethodonex{} vs 35,281.4 ms for Chodosh) on real point clouds, while being over 1000$\times$ cheaper to train compared to the cost of human annotations (\$394 vs $\sim$\$750,000).
  \item We release high quality flow pseudo-labels (representing 7.1 GPU months of compute) for the popular Argoverse~2~\citep{argoverse2} and Waymo Open~\citep{waymoopen} autonomous vehicle datasets, alongside our code and trained model weights, to facilitate further research.
\end{itemize}

\section{Background and Related Work}\sectionlabel{background}

Given point clouds $\pointcloudt{}$ at time $t$ and $\pointcloudtpone{}$ at time $t+1$, scene flow estimators predict $\flowttpone{}$, a 3D vector for each point in $\pointcloudt$ that describes how it moved from $t$ to $t+1$ \citep{dewan2016rigid}. Performance is traditionally measured using the Endpoint Error (EPE) between the predicted flow $\flowttpone{}$ and ground truth flow $\flowgtttpone{}$ (\equationref{averageepedef}):
\begin{equation}
  \small
  \equationlabel{averageepedef}
  \textup{EPE}\left({\pointcloudt} \right) = \frac{1}{\norm{\pointcloudt}} \sum_{p \in \pointcloudt} \norm{\flowttpone{}(p) - \flowgtttpone{}(p)}_2.
\end{equation}

Unlike next token prediction in language~\citep{gpt} or next frame prediction in vision~\citep{weng2021inverting}, future observations do not provide ground truth scene flow (\figref{example_sceneflow}).
To simply evaluate scene flow estimates, ground truth motion descriptions must be provided by an oracle, typically human annotation of real data \citep{waymoopen, argoverse2} or the generator of synthetic datasets \citep{flyingthings,zheng2023point}.


\begin{figure}[t]
\centering
\begin{minipage}[l]{0.28\textwidth}
\centering
\includegraphics[trim=0cm 0 0cm 0cm, clip, width=1\linewidth]{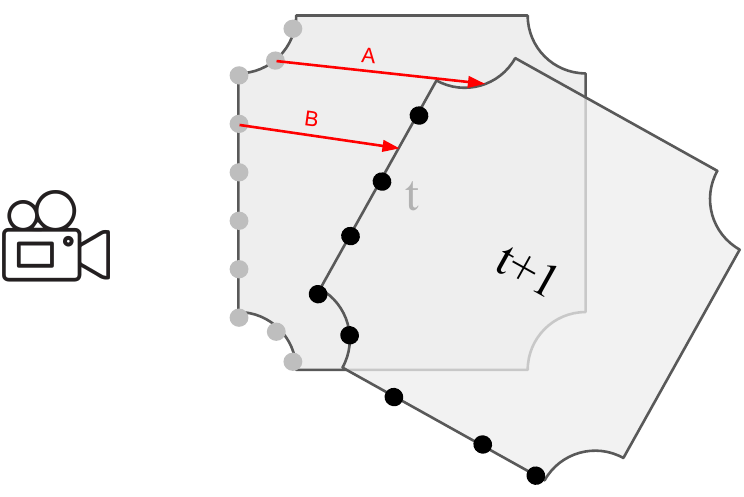}
\end{minipage} \hfill
\begin{minipage}[r]{0.67\textwidth}
\vspace{-1mm}
\hfill
    \caption{Scene Flow vectors describe where the point on an object at time $t$ will end up on the object at $t+1$. In this example, ground truth flow vector \emph{A}, associated with a point in the upper left concave corner of the object at $t$ has no nearby observations at $t+1$ due to occlusion of the concave feature. The ground truth flow vector \emph{B}, associated with a point on the face of the object at $t$, does not directly match with any observed point on the object at $t+1$ due to observational sparsity. Thus, point matching between $t$ and $t+1$ alone is insufficient to generate ground truth flow.}
    \figlabel{example_sceneflow}
\end{minipage}
\vspace{-5mm}
\end{figure}

Recent scene flow estimation methods either train feedforward methods via supervision from human annotations~\citep{flownet3d,behl2019pointflownet,tishchenko2020self,kittenplon2021flowstep3d,wu2020pointpwc,puy2020flot,li2021hcrf,scalablesceneflow,gu2019hplflownet,battrawy2022rms, 9856954}, perform human-designed test-time surrogate objective optimization over hand-designed representations~\citep{pontes2020scene,eisenberger2020smooth,nsfp,chodosh2023}, or learn from self-supervision from human-designed surrogate objectives~\citep{Mittal_2020_CVPR,baur2021slim,gojcic2021weakly,dong2022exploiting,li2022rigidflow}. 

Supervised feedforward methods are efficient at test-time; however, they require costly human annotations at train-time. Both test-time optimization and self-supervised feedforward methods seek to address this problem by optimizing or learning against label-free surrogate objectives, e.g.\ Chamfer distance~\citep{pontes2020scene}, cycle-consistency~\citep{Mittal_2020_CVPR}, and various hand-designed rigidity priors~\citep{dewan2016rigid,pontes2020scene,li2022rigidflow,chodosh2023,baur2021slim,gojcic2021weakly}. Self-supervised methods achieve faster inference by forgoing expensive test-time optimization, but do not match the quality of optimization-based methods~\citep{chodosh2023} and tend to require human-designed priors via more sophisticated network architectures compared to supervised methods~\citep{baur2021slim, gojcic2021weakly, kittenplon2021flowstep3d}. In practice, this makes them slower and more difficult to train. In contrast to existing work, we take advantage of the quality of optimization-based methods as well as the efficiency and architectural simplicity of supervised networks. Our approach, \ourmethod{}, uses label-free optimization methods~\citep{nsfp} to produce pseudo-labels to supervise a feedforward model~\citep{scalablesceneflow}, similar to methods used for distillation in other domains~\cite{yang2023adcnet}.

\section{Method}

\begin{figure}[h]
  \centering
  \input{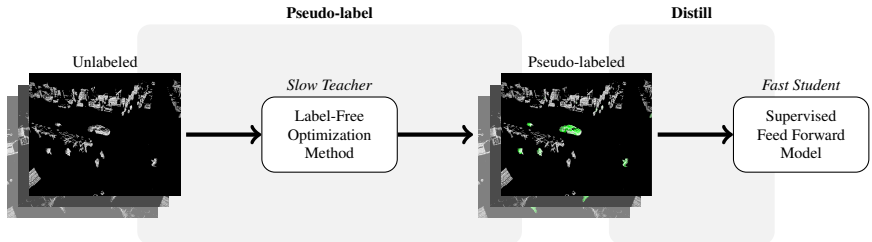}
  \caption{The \emph{\ourpipelinefull{}} (\ourpipeline{}) framework, which describes a new class of scene flow methods that produce high quality, human label-free flow at the speed of feedforward networks.}
  \figlabel{method_diagram}
\end{figure}

We propose \emph{\ourpipelinefull{}} (\ourpipeline{}), a simple, scalable distillation framework that creates a new class of scene flow estimators by using a label-free optimization method to produce pseudo-labels to supervise a feedforward model (\figref{method_diagram}). While conceptually simple, efficiently instantiating \ourpipeline{} requires careful construction; most online optimization methods and feedforward architectures are unable to efficiently scale to full-size point clouds (\sectionref{scalability}).

Based on our scalability analysis, we propose \emph{\ourmethodfull{}} (\ourmethod{}), a family of scene flow models based on \ourpipeline{} that produces fast, \textbf{state-of-the-art} scene flow estimates for full-size point clouds without any human labels (\algoref{ourmethod}). \ourmethod{} uses Neural Scene Flow prior (NSFP)~\citep{nsfp} to generate high quality, label-free pseudo-labels on full-size point clouds (\sectionref{nsfp}) and FastFlow3D~\citep{scalablesceneflow} 
for efficient inference (\sectionref{fastflow3d}).

\subsection{Scaling \ourpipelinefull{} to Large Point Clouds}\sectionlabel{scalability}

Popular AV datasets including Argoverse~2~(\cite{argoverse2}, collected with dual Velodyne VLP-32 sensors) and Waymo Open~(\cite{waymoopen}, collected with a proprietary lidar sensor and subsampled) have full-size point clouds with an average of 52,000 and 79,000 points per frame, respectively, after ground plane removal (\appendixref{datasetdetails}, \figref{pointcloud_size}). For practical applications, sensors such as the Velodyne VLP-128 in dual return mode produce up to 480,000 points per sweep~\citep{velodyne128datasheet} and proprietary sensors at full resolution can produce well over 1 million points per sweep. Thus, scene flow methods must be able to process many points in real-world applications.

Unfortunately, most existing methods focus strictly on scene flow \emph{quality} for toy-sized point clouds, constructed by randomly subsampling full point clouds down to 8,192 points~\citep{jin2022deformation,tishchenko2020self,wu2020pointpwc,kittenplon2021flowstep3d,flownet3d,nsfp}. As we are motivated by real-world applications, we instead target scene flow estimation for the full-sized point cloud, making architectural efficiency of paramount importance. As an example of stark differences between feedforward architectures, FastFlow3D \citep{scalablesceneflow}, which uses a PointPillar-style encoder \citep{pointpillars}, can process 1 million points in under 100 ms on an NVIDIA Tesla P1000 GPU (making it real-time for a 10Hz LiDAR), while methods like FlowNet3D~\citep{flownet3d}
take almost 4 seconds to process the same point cloud.

We design our approach to efficiently process full-size point clouds. For \ourpipeline{}'s pseudo-labeling step, speed is less of a concern; pseudo-labeling each point cloud pair is offline and highly parallelizable. High-quality methods like Neural Scene Flow Prior (NSFP, \cite{nsfp}) require only a modest amount of GPU memory (under $3$GB) when estimating scene flow on point clouds with 70K points, enabling fast and low-cost pseudo-labeling using a cluster of commodity GPUs; as an example, pseudo-labeling the Argoverse 2 train split with NSFP is over 1000$\times$ cheaper than human annotation (\appendixref{labelvspseudolabelcosts}). The efficiency of \ourpipeline{}'s student feedforward model \emph{is} critical, as it determines both the method's test-time speed and its training speed (faster training enables scaling to larger datasets), motivating models that can efficiently process full-size point clouds.

\subsection{Neural Scene Flow Prior is a Slow Teacher}\sectionlabel{nsfp}

Neural Scene Flow Prior (NSFP, \cite{nsfp}) is an optimization-based approach to scene flow estimation. Notably, it does not use ground truth labels to generate high quality flows, instead relying upon strong priors in its learnable function class (determined by the coordinate network's architecture) and optimization objective (\equationref{nsfploss}). Point residuals are fit per point cloud pair $\pointcloudt$, $\pointcloudtpone{}$ at test-time by randomly initializing two MLPs; one to describe the forward flow $\flowforward$ from $\pointcloudt$ to $\pointcloudtpone{}$, and one to describe the reverse flow $\flowrev$ from $\pointcloudt{} + \flowttpone$ to $\pointcloudt{}$ in order to impose cycle consistency. The forward flow $\flowforward$ and backward flow $\flowrev$ are optimized jointly to minimize
\begin{equation}
\small
  \equationlabel{nsfploss}
  \chamferdistance{\pointcloudt{} + \flowforward}{\pointcloudtpone} + \chamferdistance{\pointcloudt{} + \flowforward + \flowrev}{\pointcloudt} \enspace ,
\end{equation}
where $\chamferdistancename$ is the standard Chamfer distance with per-point distances above 2 meters set to zero to reduce the influence of outliers.

NSFP is able to produce high-quality scene flow estimations due to its choice of coordinate network architecture and use of cycle consistency constraint. The coordinate network's learnable function class is expressive enough to fit the low frequency signal of residuals for moving objects while restrictive enough to avoid fitting the high frequency noise from $\chamferdistancename$, and the cycle consistency constraint acts as a local smoothness regularizer for the forward flow, as any shattering effects in the forward flow are penalized by the backwards flow. NSFP provides high quality estimates on full-size point clouds (\figref{tradeoff_curve}), so we select NSFP for \ourmethod{}'s pseudo-label step of \ourpipeline{}.

\subsection{FastFlow3D is a Fast Student}\sectionlabel{fastflow3d}
FastFlow3D~\citep{scalablesceneflow} is an efficient feedforward method that learns using human supervisory labels $\flowgtttpone{}$ and per-point foreground / background class labels. FastFlow3D's loss minimizes a variation of the End-Point Error (\equationref{averageepedef}) that reduces the importance of annotated background points, thus minimizing

\vspace{-1.2em}
\begin{tabular}{ccc}
\small
\begin{minipage}[b]{0.44\textwidth}
\begin{equation}
  \equationlabel{fastflowloss}
  \!\frac{1}{\norm{\pointcloudt}} \sum_{p \in \pointcloudt} \!\bgscale{p} \norm{\flowttpone{}(p) - \flowgtttpone{}(p)}_2
\end{equation}
\end{minipage}   & where &
\begin{minipage}[b]{0.42\textwidth}
\begin{equation}
  \equationlabel{bgscale}
  \bgscale{p} = \begin{cases}
    1   & \text{if } p \in \text{Foreground} \\
    0.1 & \text{if } p \in \text{Background} \enspace .
  \end{cases}
\end{equation}
\end{minipage}
\end{tabular}

FastFlow3D's architecture is a PointPillars-style encoder~\citep{pointpillars}, traditionally used for efficient LiDAR object detection \citep{ Vedder2022sparsepointpillars}, that converts the point cloud into a birds-eye-view pseudoimage using infinitely tall voxels (pillars). This pseudoimage is then processed with a 4 layer U-Net style backbone. The encoder of the U-Net processes the $\pointcloudt$ and $\pointcloudtpone$ pseudoimage separately, and the decoder jointly processes both pseudoimages. A small MLP is used to decode flow for each point in $\pointcloudt{}$ using the point's coordinate and its associated pseudoimage feature.

As discussed in \sectionref{scalability}, FastFlow3D's architectural design choices make fast even on full-size point clouds. While most feedforward methods are evaluated using a standard toy evaluation protocol with subsampled point clouds, FastFlow3D is able to scale up to full resolution point clouds while maintaining real-time performance and emitting competitive quality scene flow estimates using human supervision, making it a good candidate for the distillation step of \ourpipeline{}. 

In order to train FastFlow3D using pseudo-labels, we replace the foreground / background scaling function (\equationref{bgscale}) with a simple uniform weighting ($\bgscale{\cdot} = 1$), which collapses to Average EPE; see \appendixref{speedscalingexperiments} for experiments with other weighting schemes. Additionally, we depart from FastFlow3D's problem setup in two minor ways: we delete ground points using dataset provided maps, a standard pre-processing step \citep{chodosh2023}, and use the standard scene flow problem setup of predicting flow between two frames (\sectionref{background}) instead of predicting future flow vectors in meters per second. \algoref{ourmethod} describes our approach, with details specified in \sectionref{methodperf}.

In order to take advantage of the unlabeled data scaling of \ourpipeline{}, we expand FastFlow3D to a family of models by designing a higher capacity backbone, producing \emph{FastFlow3D XL}. This larger backbone halves the size of each pillar to quadruple the pseudoimage area, doubles the size of the pillar embedding, and adds an additional layer to maintain the network's receptive field in metric space; as a result, the total parameter count increases from 6.8 million to 110 million.
\vspace{-0.25em}
\begin{algorithm}
\caption{\ourmethod{}}\algolabel{ourmethod}
\begin{algorithmic}[1]
\State $D \gets $ collection of unlabeled point cloud pairs \Comment{Training Data}
\For{$\pointcloudt, \pointcloudtpone \in D$}\Comment{Parallel \texttt{For}}
\State $\flowgtttpone \gets \textup{Teacher{NSFP}}(\pointcloudt, \pointcloudtpone)$ \Comment{\ourpipeline{} \emph{Pseudo-label} Step}
\EndFor
\For {epoch $\in$ epochs} 
\For{$\pointcloudt, \pointcloudtpone, \flowgtttpone \in D$} \Comment{\ourpipeline{}'s \emph{Distill} Step}
\State $l \gets \textup{\equationref{fastflowloss}}(\textup{StudentFastFlow3D}_\theta(\pointcloudt, \pointcloudtpone), \flowgtttpone)$
\State $\theta \gets \theta$ updated w.r.t.\ $l$
\EndFor
\EndFor
\end{algorithmic}
\end{algorithm}
\vspace{-1.75em}
\section{Experiments}\sectionlabel{experiments}

\ourmethod{} provides a family of fast, high quality scene flow estimators. In order to validate this family and understand the impact of components in the underlying \ourpipelinefull{} framework, we perform extensive experiments on the Argoverse~2~\citep{argoverse2} and Waymo Open~\citep{waymoopen} datasets. We compare to author implementations of NSFP~\citep{nsfp} and ~\citet{chodosh2023}, implement FastFlow3D~\citep{scalablesceneflow} ourselves (no author implementation is available), and use \citet{chodosh2023}'s implementations for all other baselines.

As discussed in \citet{chodosh2023}, downstream applications typically rely on good quality scene flow estimates for foreground points. Most scene flow methods are evaluated using average Endpoint Error (EPE, \equationref{averageepedef});  however, roughly 80\% of real-world point clouds are background, causing average EPE to be dominated by background point performance. To address this, we use the improved evaluation metric proposed by \citet{chodosh2023}, \emph{Threeway EPE}:
\begin{equation}
  \small
  \equationlabel{threewayepedef}
  \textup{Threeway EPE}(\pointcloudt) = \textup{Avg} \begin{cases}
      \textup{EPE}({p \in \pointcloudt:  p \in \textup{Background}}) & \textup{(Static BG)} \\
      \textup{EPE}({p \in \pointcloudt: p \in \textup{Foreground}\land \flowgtttpone{}(p) \leq 0.5 \textup{m/s}})  & \textup{(Static FG)} \\
      \textup{EPE}({p \in \pointcloudt: p \in \textup{Foreground}\land \flowgtttpone{}(p) > 0.5 \textup{m/s}})  & \textup{(Dynamic FG)}  \enspace .\\
  \end{cases}
\end{equation}

\subsection{How does \ourmethod{} perform compared to prior art on real point clouds?}\sectionlabel{methodperf}

The overarching promise of \ourmethod{} is the ability to build fast, high quality scene flow estimators that improve with the the availability of large-scale \emph{unlabeled} data. Does \ourmethod{} deliver on this promise? How does it compare to state-of-the-art methods?

To characterize the \ourmethod{} family's performance, we use Argoverse~2 to perform scaling experiments along two axes: dataset size and student size. For our standard size configuration, we use the Argoverse~2 Sensor \emph{train} split and the standard FastFlow3D architecture, enabling head-to-head comparisons against the fully supervised FastFlow3D as well as other baseline methods. For our scaled up dataset (denoted \emph{\threex} in all experiments), we use the Argoverse~2 Sensor \emph{train} split and concatenate a roughly twice as large set of unannotated frame pairs from the Argoverse~2 LiDAR dataset, uniformly sampled from its 20,000 sequences to maximize data diversity. For our scaled up student architecture (denoted \emph{\xl} in all experiments), we use the \xl{} backbone described in \sectionref{fastflow3d}. For details on the exact dataset construction and method hyperparameters, see \appendixref{datasetdetails}

\begin{table}[h]
  \caption{Quantitative results on the Argoverse~2 Sensor validation split using the evaluation protocol from \citet{chodosh2023}. The methods used in this paper, shown in the first two blocks of the table, are trained and evaluated on point clouds within a 102.4m $\times$ 102.4m area centered around the ego vehicle (the settings for the \emph{Argoverse~2 Self-Supervised Scene Flow Challenge}) . However, following the protocol of \citet{chodosh2023}, all methods report error on points in the 70m $\times$ 70m area centered around the ego vehicle. Runtimes are collected on an NVIDIA V100 with a batch size of 1 \citep{peri2023empirical}. FastFlow3D, \ourmethodonex{}, and \ourmethodthreex{} have identical feedforward architectures and thus share the same real-time runtime; FastFlow3D XL, \ourmethodxlonex{}, and \ourmethodxlthreex{} have identical feedforward architectures and thus share the same runtime.  Methods with an * have performance averaged over 3 training runs (see \appendixref{intertrainvariance} for details). Underlined methods require human supervision.}
  \centering
  \resulttablefontsize
  \setlength{\tabcolsep}{2.5pt}
  \begin{tabular}{lrr|r|rrrr}
 \toprule
 & \multicolumn{2}{c|}{Runtime (ms)} & Point Cloud     & Threeway & Dynamic & Static & Static\\
 & &                                 & Subsampled Size & EPE      & FG EPE  & FG EPE & BG EPE\\ \midrule
 \humanlabels{FastFlow3D}*~\citep{scalablesceneflow} & \multirow{3}{*}{${29.33\pm}$} & \multirow{3}{*}{${2.38}$} & Full Point Cloud & 0.071 & 0.186 & 0.021 & 0.006 \\
 \ourmethodonex{}* (Ours) & & & Full Point Cloud & 0.088 & 0.231 & 0.022 & 0.011 \\
  \ourmethodthreex{} (Ours) & & & Full Point Cloud & 0.064 & 0.164 & 0.017 & 0.011  \\
  \ourmethodfivex{} (Ours) & & & Full Point Cloud & \textbf{0.056} & 0.140 & 0.017 & 0.011  \\ \midrule
  \humanlabels{FastFlow3D XL}  & \multirow{3}{*}{${260.61\pm}$} & \multirow{3}{*}{${1.21}$} & Full Point Cloud & 0.055 & 0.139 & 0.018 & 0.007 \\
  \ourmethodxlonex{} (Ours) & & & Full Point Cloud & 0.070 & 0.178 & 0.019 & 0.013 \\
  \ourmethodxlthreex{} (Ours) & & & Full Point Cloud & \textbf{0.054} & 0.131 & 0.018 & 0.012 \\
 NSFP w/ Motion Comp~\citep{nsfp} & $26,285.0\pm$ & $18,139.3$ & Full Point Cloud & 0.067 & 0.131 & 0.036 & 0.034  \\
 Chodosh et al.~\citep{chodosh2023} & $35,281.4\pm$ & $20,247.7$ & Full Point Cloud & 0.055 & 0.129 & 0.028 & 0.008 \\ \midrule
 Odometry & \multicolumn{2}{c|}{---} & Full Point Cloud & 0.198 & 0.583 & 0.010 & 0.000  \\
 ICP~\citep{chen1992object} & $523.11\pm$ & $169.34$ & Full Point Cloud & 0.204 & 0.557 & 0.025 & 0.028  \\
 \humanlabels{Gojcic}~\citep{gojcic2021weakly} & $6,087.87\pm$ & $1,690.56$ & $20000$ & 0.083 & 0.155 & 0.064 & 0.032  \\
 Sim2Real~\citep{jin2022deformation} & $99.35\pm$ & $13.88$ & $8192$ & 0.157 & 0.229 & 0.106 & 0.137  \\
 EgoFlow~\citep{tishchenko2020self} & $2,116.34\pm$ & $292.32$ & $8192$ & 0.205 & 0.447 & 0.079 & 0.090  \\
 PPWC~\citep{wu2020pointpwc} & $79.43\pm$ & $2.20$ & $8192$ & 0.130 & 0.168 & 0.092 & 0.129  \\
 FlowStep3D~\citep{kittenplon2021flowstep3d} & $687.54\pm$ & $3.13$ & $8192$ & 0.161 & 0.173 & 0.132 & 0.176  \\
 \bottomrule
\end{tabular}
  \tablelabel{argobigtable}
\end{table}

As shown in \tableref{argobigtable}, \ourmethod{} is able to leverage scale to deliver superior performance. While \ourmethodonex{} loses a head-to-head competition against the human-supervised FastFlow3D on both Argoverse~2 (\tableref{argobigtable}) and Waymo Open (\tableref{waymobigtable}), scaling the distillation process to additional unlabeled data provided by Argoverse 2 enables \ourmethodthreex{} to significantly surpass the performance of both methods just by training on more pseudo-labled data. \ourmethodthreex{} even surpasses the performance of its own teacher, NSFP, \emph{while running in real-time!} 

\ourmethod{}'s pipeline also benefits from scaling up the student architecture. We modify \ourmethod{}'s architecture with the much larger \xl{} backbone, and show that our \ourmethodxlthreex{} is able to combine the power of dataset and model scale to outperform all other methods, including significantly outperform its own teacher. Our simple approach achieves \textbf{state-of-the-art} on both the Argoverse~2 validation split and \textbf{\emph{Argoverse~2 Self-Supervised Scene Flow Challenge}}.




\begin{table}[h]
  \centering
  \caption{Quantitative results on Waymo Open using the evaluation protocol from \citet{chodosh2023}. Runtimes are scaled to approximate the performance on a V100~\citep{li2020towards}. Both FastFlow3D and \ourmethodonex{} have identical feedforward architectures and thus share the same runtime. Underlined methods require human supervision.}
  \resulttablefontsize
  \setlength{\tabcolsep}{3pt}
  \begin{tabular}{lrr|r|rrrr}
 \toprule
 & \multicolumn{2}{c|}{Runtime (ms)} & Point Cloud     & Threeway & Dynamic & Static & Static\\
 & &                                 & Subsampled Size & EPE      & FG EPE  & FG EPE & BG EPE\\ \midrule
 \ourmethodonex{} (Ours) & \multirow{2}{*}{$21.66\pm$} & \multirow{2}{*}{$0.48$} & Full Point Cloud & 0.092 & 0.216 & 0.015 & 0.045 \\
 \humanlabels{FastFlow3D}~\citep{scalablesceneflow} & & & Full Point Cloud & 0.078 & 0.195 & 0.015 & 0.024 \\
 \midrule
 Chodosh~\citep{chodosh2023} & $93,752.3\pm$ & $76,786.1$ & Full Point Cloud & \textbf{0.041} & 0.073 & 0.013 & 0.039 \\
 NSFP~\cite{nsfp} & $90,999.1\pm$ & $74,034.9$ & Full Point Cloud & 0.100 & 0.171 & 0.022 & 0.108 \\
 ICP~\citep{chen1992object} & $302.70\pm$ & $157.61$ & Full Point Cloud & 0.192 & 0.498 & 0.022 & 0.055 \\
 \humanlabels{Gojcic}~\cite{gojcic2021weakly} & $501.69\pm$ & $54.63$ & 20000 & 0.059 & 0.107 & 0.045 & 0.025 \\
 EgoFlow~\citep{tishchenko2020self} & $893.68\pm$ & $86.55$ & 8192 & 0.183 & 0.390 & 0.069 & 0.089 \\
 Sim2Real~\citep{jin2022deformation} & $72.84\pm$ & $14.79$ & 8192 & 0.166 & 0.198 & 0.099 & 0.201 \\
 PPWC~\citep{wu2020pointpwc} & $101.43\pm$ & $5.48$ & 8192 & 0.132 & 0.180 & 0.075 & 0.142 \\
 FlowStep3D~\citep{kittenplon2021flowstep3d} & $872.02\pm$ & $6.24$ & 8192 & 0.169 & 0.152 & 0.123 & 0.232\\
 \bottomrule
\end{tabular} 
  \tablelabel{waymobigtable}
\end{table}

\subsection{How does \ourmethod{} scale?}\sectionlabel{scalinglaws}

\sectionref{methodperf} demonstrates that \ourmethod{} can leverage scale to capture state-of-the-art performance. However, it's difficult to perform extensive model tuning for large training runs, so predictable estimates of performance as a function of dataset size are critical \citep{openai2023gpt4}. 
Does \ourmethod{}'s performance follow predictable scaling laws?

We train \ourmethod{} and FastFlow3D on sequence subsets / supersets of the Argoverse~2 Sensor train split. \figref{scaling_log} shows \ourmethod{} and FastFlow3D's validation Threeway EPE both decrease roughly logarithmically, and this trend appears to hold for \xl{} backbone models as well.

\begin{figure}[htb]
  \includegraphics{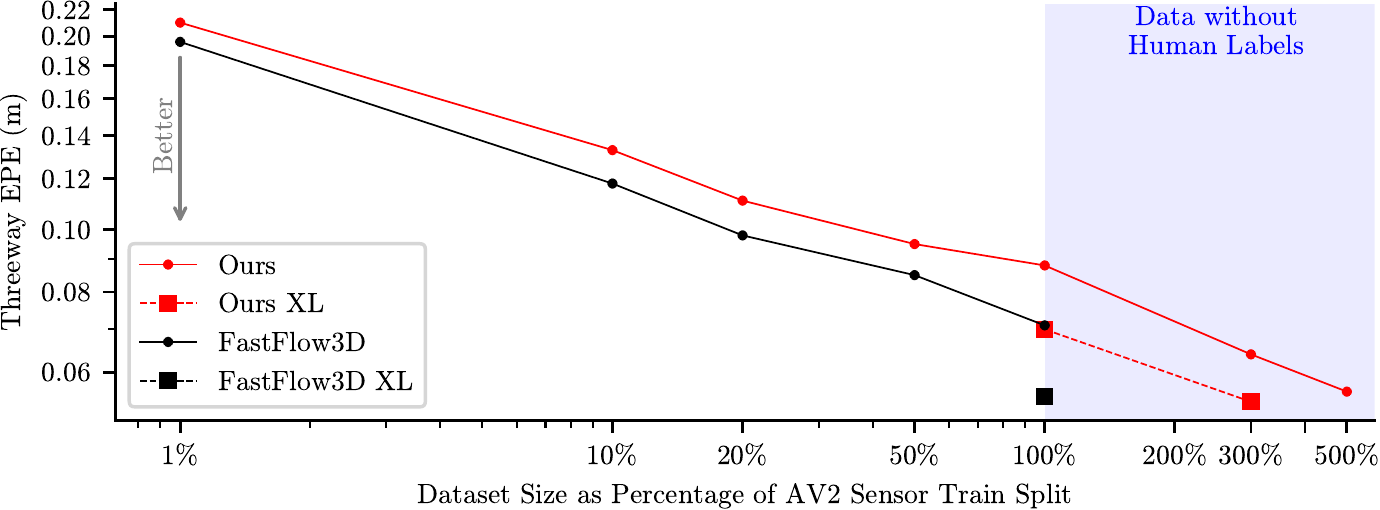}
\caption{Empirical scaling laws for \ourmethod{}. We report Argoverse 2 validation split Threeway EPE as a percentage of the Argoverse 2 \emph{train} split used, on a log$_{10}$-log$_{10}$ scale, trained to convergence. Threeway EPE performance of \ourmethod{} scales logarithmically with the amount of  training data.}
\figlabel{scaling_log}
\end{figure}

Empirically, \ourmethod{} adheres to predictable scaling laws that demonstrate more data (and more parameters) are all you need to get better performance. This makes \ourmethod{} a practical pipeline for building \emph{scene flow foundation models}~\citep{Bommasani2021FoundationModels} using the raw point cloud data that exists \emph{today} in the deployment logs of Autonomous Vehicles and other deployed systems.

\subsection{How does dataset diversity influence \ourmethod{}'s performance?}

In typical human annotation setups, a point cloud \emph{sequence} is given to the human annotator. The human generates box annotations in the first frame, and then updates the pose of those boxes as the objects move through the sequence, introducing and removing annotations as needed. This process is much more efficient than annotating disjoint frame pairs, as it amortizes the time spent annotating most objects in the sequence. This is why most human annotated training datasets (e.g. Argoverse~2 Sensor, Waymo Open) are composed of contiguous \emph{sequences}.
However, contiguous frames have significant structural similarity; in the 150 frames (15 seconds) of an Argoverse~2 Sensor sequence, the vehicle typically observes no more than a city block's worth of unique structure. 
\ourmethod{}, which requires \emph{zero} human labels, does not have this constraint on its pseudo-labels; NSFP run on non-sequential frames is no more expensive than NSFP run on non-sequential frames, enabling \ourmethod{} to train on a more diverse dataset. 
How does dataset diversity impact performance?

To understand the impact of data diversity, we train a version of \ourmethodonex{} and \ourmethodtwox{} \emph{only} on the diverse subset of our Argoverse~2 LiDAR data selected by uniformly sampling 12 frame pairs from each of the 20,000 unique sequences (\tableref{argodiversity}). 

\begin{table}[h]
  \centering
  \caption{Comparison between \ourmethod{} trained on Argoverse~2 Sensor dataset versus the more diverse, unlabeled Argoverse~2 LiDAR subset described in \sectionref{methodperf}. Diverse training datasets result in non-trivial performance improvements.}
  \scriptsize
  \setlength{\tabcolsep}{3pt}
  \begin{tabular}{lrrrr}
\toprule
& Threeway & Dynamic & Static & Static  \\
& EPE & FG EPE & FG EPE & BG EPE \\
\midrule
\humanlabels{FastFlow3D}*~\citep{scalablesceneflow} & $0.071$ & $0.186$ & $0.021$ & $0.006$  \\
\ourmethodonex{}  (AV2 Sensor Data)*
& $0.088$ & $0.231$ & $0.022$ & $0.011$  \\
\ourmethodonex{} (AV2 LiDAR Subset Data) 
& $0.082$ & $0.218$ & $0.018$ & $0.009$ \\
\ourmethodtwox{} (AV2 LiDAR Subset Data) 
& $0.072$ & $0.184$ & $0.022$ & $0.011$ \\
\bottomrule
\end{tabular}
  \tablelabel{argodiversity}
\end{table}

Dataset diversity has a non-trivial impact on performance; \ourmethod{}, by virtue of being able to learn across \emph{non-contiguous} frame pairs, is able to see more unique scene structure and thus learn to better to extract motion in the presence of the unique geometries of the real world.

 \subsection{How do the noise characteristics of \ourmethod{} compare to other methods?}\sectionlabel{noise_of_our_method}

 \ourmethod{} distills NSFP into a feedforward model from the FastFlow3D family. \sectionref{methodperf} highlights the \emph{average} performance of \ourmethod{} across Threeway EPE catagories, but what does the error \emph{distribution} look like?

 \begin{figure}[h!]
  \figlabel{endpoint_distribution}
  \centering
  \input{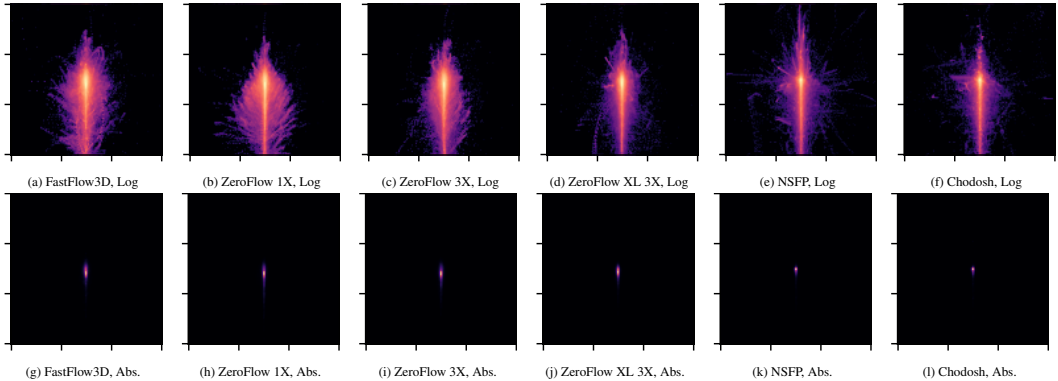}
  \caption{Normalized frame birds-eye-view heatmaps of endpoint residuals for Chamfer Distance, as well as the outputs for NSFP and Chodosh on moving points (points with ground truth speed above 0.5m/s). Perfect predictions would produce a single central dot. Top row shows the frequency on a $\log_{10}$ color scale, bottom row shows the frequency on an absolute color scale. Qualitatively, methods with better quantitative results have tighter residual distributions. See \appendixref{endpoint_errors_details} for details.}
  \figlabel{endpoint_distribution_test_time}
\end{figure}

 To answer this question, we plot birds-eye-view flow vector residuals of NSFP, Chodosh, FastFlow3D, and several members of the \ourmethod{} family on moving objects from the Argoverse~2 validation dataset, where the ground truth is rotated vertically and centered at the origin to present all vectors in the same frame (\figref{endpoint_distribution_test_time}; see \appendixref{endpoint_errors_details} for details on construction).
Qualitatively, these plots show that error is mostly distributed along the camera ray and distributional tightness ($\log_{10}$ plots) roughly corresponds to overall method performance.

Overall, these plots provide useful insights to practitioners and researchers, particularly for consumption in downstream tasks; as an example, open world object extraction~\citep{objectdetectionmotion} requires the ability to threshold for motion and cluster motion vectors together to extract the entire object. Decreased average EPE is useful for this task, but understanding the magnitude and \emph{distribution} of flow vectors is needed to craft good extraction heuristics.


\subsection{How does teacher quality impact \ourmethod{}'s performance?}\sectionlabel{chodoshpseudolabels}

As shown in \sectionref{methodperf} ~\citep{chodosh2023} has superior Threeway EPE over NSFP on both Argoverse~2 and Waymo Open. Can a better performing teacher lead a better version of \ourmethod{}?

To understand the impact of a better teacher, we train \ourmethod{} on Argoverse 2 using superior quality flow vectors from \citet{chodosh2023}, which proposes a refinement step to NSFP lablels to provide improvements to flow vector quality (\tableref{argochodosh}). \ourmethod{} trained on Chodosh refined pseudo-labels provides no meaningful quality improvement over NSFP pseudo-labels (as discussed in \appendixref{intertrainvariance}, the Threeway EPE difference is within training variance for \ourmethod{}). These results also hold for our ablated speed scaled version of \ourmethod{} in \appendixref{speedscalingexperiments}.

Since increasing the quality of the teacher over NSFP provides no noticeable benefit, can we get away with using a significantly faster but lower quality teacher to replace NSFP, e.g.\ the commonly used self-supervised proxy of \chamferdistancenameraw{}? 

To understand if NSFP is necessary, we train ZeroFlow on Argoverse~2 using pseudo-labels from the nearest neighbor, truncated to 2 meters as with \chamferdistancenameraw{}. The residual distribution of \chamferdistancenameraw{} is shown in \appendixref{endpoint_errors_details}, \figref{endpoint_distribution_nearest_neighbor}. \ourmethod{} trained on \chamferdistancenameraw{} pseudo-labels performs significantly worse than NSFP, motivating the use of NSFP as a teacher.
 
\begin{table}[h]
  \centering
  \caption{Comparison between \ourmethod{} trained on Argoverse 2 using NSFP pseudo-labels, \ourmethod{} using \citet{chodosh2023} pseudo-labels, and \ourmethod{} using \chamferdistancenameraw{}. Methods with an * have performance averaged over 3 training runs (see \appendixref{intertrainvariance} for details). The minor quality improvement of Chodosh pseudo-labels does not lead to a meaningful difference in performance, while the significant degradation of \chamferdistancenameraw{} leads to significantly worse performance.}
  \scriptsize
  \setlength{\tabcolsep}{3pt}
  \begin{tabular}{lrrrr}
\toprule
& Threeway & Dynamic & Static & Static \\
& EPE & FG EPE & FG EPE & BG EPE  \\
\midrule
\ourmethodonex{}  (NSFP pseudo-labels)* 
& $0.088$ & $0.231$ & $0.022$ & $0.011$  \\
\ourmethodonex{} (\citet{chodosh2023} pseudo-labels) 
& $0.085$ & $0.234$ & $0.018$ & $0.004$ \\
\ourmethodonex{} (\chamferdistancenameraw{} pseudo-labels) 
& $0.105$ & $0.226$ & $0.049$ & $0.040$ \\
\bottomrule
\end{tabular}
  \tablelabel{argochodosh}
\end{table}

\section{Conclusion}\sectionlabel{conclusion}

Our scene flow approach, \ourmethodfull{} (\ourmethod{}), produces fast, state-of-the-art scene flow \emph{without human labels} via our conceptually simple distillation pipeline. 

But, more importantly, we present the first  practical pipeline for building \emph{scene flow foundation models}~\citep{Bommasani2021FoundationModels} using the raw point cloud data that exists \emph{today} in the deployment logs at Autonomous Vehicle companies and other deployed robotics systems. Foundational models in other domains like language~\citep{gpt3,openai2023gpt4} and vision~\citep{segmentanything,r3m} have enabled significant system capabilities with little or no additional domain-specific fine-tuning~\citep{wang2023voyager,ma2022vip,ma2023liv}. We posit that a scene flow foundational model will enable new systems that can leverage high quality, general scene flow estimates to robustly reason about object dynamics even in foreign or noisy environments.

\poorparagraph{Limitations and Future Work} \ourmethod{} inherits the biases of its pseudo-labels. Unsurprisingly, if the pseudo-labels consistently fail to estimate scene flow for certian objects, our method will also be unable to predict scene flow for those objects; however, further innovation in model architecture, loss functions, and pseudo-labels may yield better performance. In order to enable further work on \ourpipelinefull{}-based methods, we release\footnote{\url{https://vedder.io/zeroflow}} our code, trained model weights, and NSFP flow pseudo-labels, representing $3.6$ GPU months for Argoverse 2 and $3.5$ GPU months for Waymo Open.

\poorparagraph{Acknowledgements} The research presented in this paper was partially supported by the DARPA SAIL-ON program under contract HR001120C0040, the DARPA ShELL program under agreement HR00112190133, the Army Research Office under MURI grant W911NF20-1-0080, and the CMU Center for Autonomous Vehicle Research. This work was performed under the auspices of the U.S. Department of Energy by Lawrence Livermore National Laboratory under Contract DE-AC52-07NA27344.

\bibliography{references}
\bibliographystyle{iclr2024_conference}

\newpage 

\appendix
\section{Argoverse 2 and Waymo Open Dataset Configuration Details}\appendixlabel{datasetdetails}

\poorparagraph{Argoverse 2} The Sensor dataset contains 700 training and 150 validation sequences. Each sequence contains 15 seconds of 10Hz point clouds collected using two Velodyne VLP-32s mounted on the roof of a car.
As part of the training protocol for \ourmethod{}, FastFlow3D, and NSFP w/ Motion Compensation, we perform ego compensation, ground point removal, and restrict all points to be within a 102.4m $\times$ 102.4m area centered around the ego vehicle, resulting in point clouds with an average of 52,871 points (\figref{argo_pointcloud_size}). The point cloud $\pointcloudtpone{}$ is centered at the origin of the ego vehicle's coordinate system and $\pointcloudt{}$ is projected into $\pointcloudtpone{}$'s coordinate frame. For \ourmethod{} and FastFlow3D, the PointPillars encoder uses $0.2$m$\times 0.2$m pillars, with all architectural configurations matching \citep{scalablesceneflow}. For NSFP w/ Motion Compensation, we use the same architecture and early stopping parameters as the original method~\citep{nsfp}. For FastFlow3D and the FastFlow3D student architecture of \ourmethod{}, we train to convergence (50 epochs) with an Adam~\citep{kingma2014adam} learning rate of \SI{2e-6} and batch size $64$. For FastFlow3D XL and the FastFlow3D XL student architecture of \ourmethod{} (\ourmethodxlonex{}, \ourmethodxlthreex{}), we train to convergence (10 epochs) with the same optimizer settings and a batch size $12$. For \ourmethodthreex{} and and \ourmethodxlthreex{}, we train on an additional 240,000 unlabeled frame pairs (roughly twice the size as the Argoverse~2 Sensor \emph{train} split), constructed by selecting 12 frame pairs at uniform intervals from the 20,000 sequences of the Argoverse~2 LiDAR dataset. For all other methods in \tableref{argobigtable}, we use the implementations provided by \citet{chodosh2023}, which follow ground removal and ego compensation protocols from their respective papers.

\poorparagraph{Waymo Open} The dataset contains 798 training and 202 validation sequences. Each sequence contains 20 seconds of 10Hz point clouds collected using a custom LiDAR mounted on the roof of a car.
We use the same preprocessing and training configurations used on Argoverse 2; after ego motion compensation and ground point removal, the average point cloud has 79,327 points (\figref{waymo_pointcloud_size}).

As shown in \figref{pointcloud_size}, Argoverse 2~\citep{argoverse2} and Waymo Open~\citep{waymoopen} are significantly larger than the 8,192 point subsampled point clouds used by prior art.

\begin{figure}[h]
  \begin{subfigure}[T]{0.48\textwidth}
    \centering
    \includegraphics{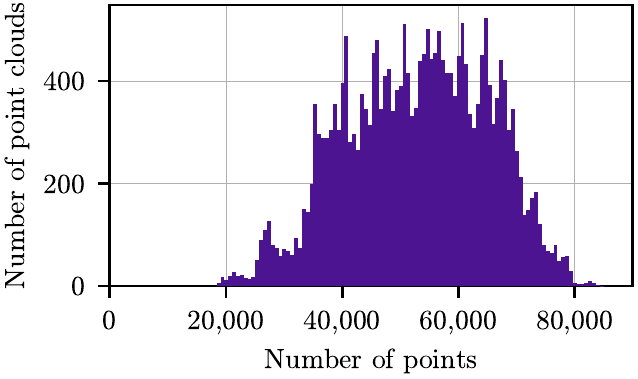}
    \caption{Distribution of point cloud sizes in the Argoverse~2 Sensor \emph{val} split: $\mu = 52,871.6; \sigma = 12,227.2$.}
    \figlabel{argo_pointcloud_size}
  \end{subfigure}
  \hfill
  \begin{subfigure}[T]{0.48\textwidth}
    \centering
    \includegraphics{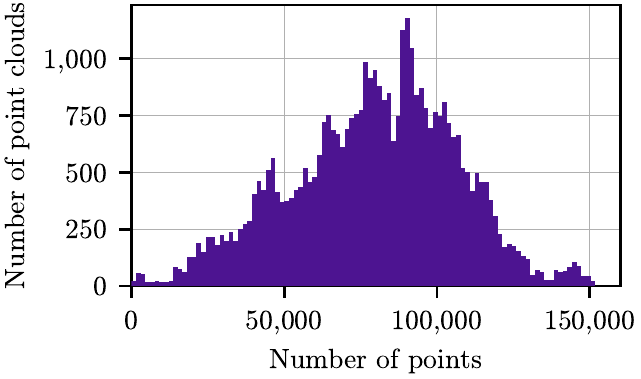}
    \caption{Distribution of point cloud sizes in the Waymo Open \emph{val} split: $\mu = 79,327.8; \sigma = 27,182.1$.}
    \figlabel{waymo_pointcloud_size}
  \end{subfigure}
  \caption{Point cloud size distributions for the \emph{val} set of the Argoverse~2 Sensor~\citep{argoverse2} and Waymo Open~\citep{waymoopen} datasets after ground removal and clipped to a 102.4m $\times$ 102.4m box around the ego vehicle.}
  \figlabel{pointcloud_size}
\end{figure}

\begin{figure}[t]
  \centering
  \begin{tikzpicture}
    \setlength{\fboxsep}{0.0pt}
    \node[anchor=south west,inner sep=0] (image) at (0,0) {\fbox{\includegraphics[width=0.47\textwidth]{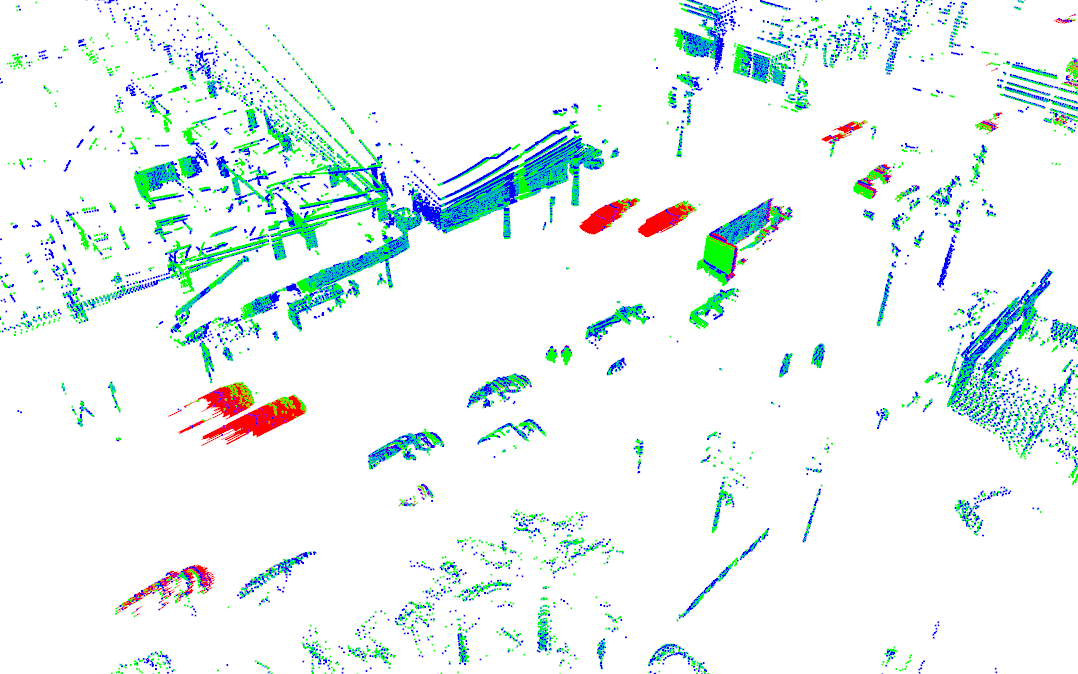}}};
    \node[anchor=south west,inner sep=0] (image2) at (6.9,0) {\fbox{\includegraphics[width=0.47\textwidth]{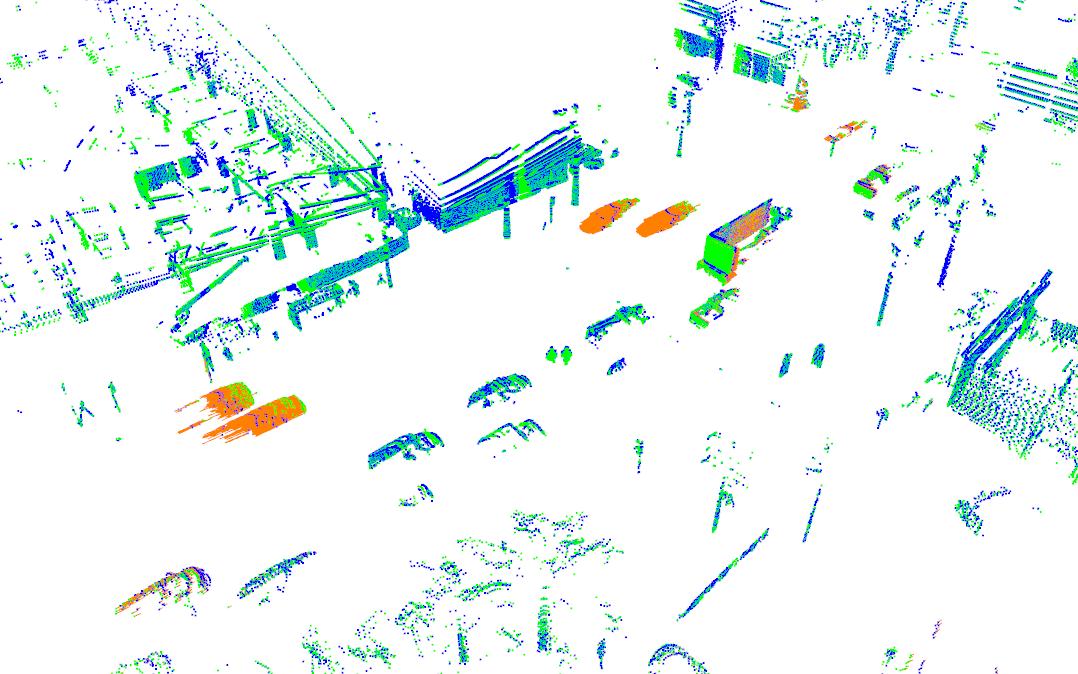}}};
    \begin{scope}[x={(image.south east)},y={(image.north west)}]
      \coordinate (zoom_area_south_west) at (0.52,0.63);
      \coordinate (zoom_area_north_east) at (0.66,0.75);
      \coordinate (zoom_area_north_west) at (0.52,0.75);

      \draw[black] (zoom_area_south_west) rectangle (zoom_area_north_east);

      \node[inner sep=0,right=-1.6cm of image, yshift=0.8cm] (zoomed_image) {{\fbox{\includegraphics[width=3.2cm]{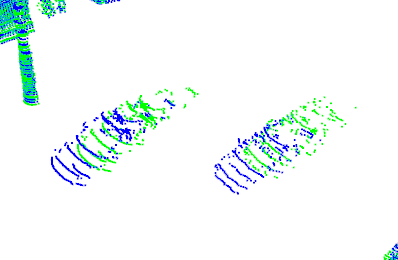}}}};

      \draw[black] (zoom_area_south_west) -- (zoomed_image.south west);
      \draw[black] (zoom_area_north_west) -- (zoomed_image.north west);
    \end{scope}
  \end{tikzpicture}
  \caption{Scene flow estimation of two consecutive point clouds sampled 100 ms apart (green and blue, respectively) on Argoverse 2~\citep{argoverse2}. \textbf{Left:} Ground truth scene flow annotations in red. These annotations are derived from the motion of amodal bounding boxes. \textbf{Right:} \ourmethod{}'s scene flow estimates estimates in orange, which closely match with the ground truth.
  }
  \figlabel{sceneflowexample}
\end{figure}
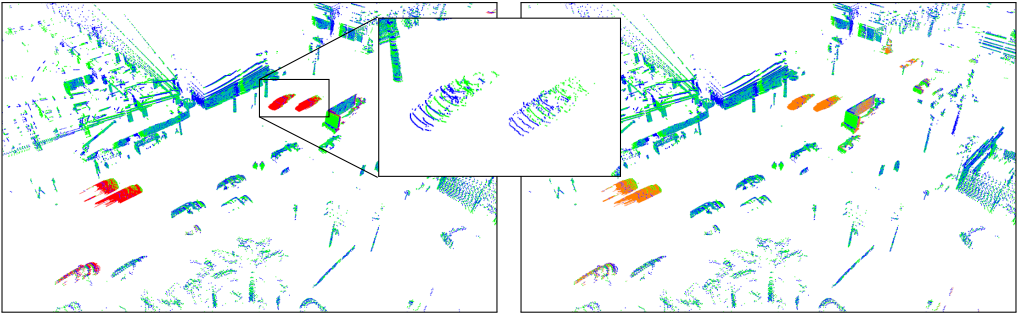

\section{Exploring the importance of point weighting}\appendixlabel{speedscalingexperiments}

In order to train FastFlow3D using pseudo-labels, we need a replacement $\bgscale{\cdot}$ semantics scaling function described in \equationref{bgscale}) because our pseudo-labels do not provide foreground / background semantics. In the main experiments, we use uniform scaling ($\bgscale{\cdot} = 1$).

\subsection{Can we design a better point weighting function for pseudo-labels?}\appendixlabel{speedscaling}

We propose a soft weighting based on pseudo-label flow magnitude: for the point $p$ in the pseudo-label flow $\flowgtttpone{}(p)$, where $\pointspeed{p}$ represents its speed in meters per second, we linearly interpolate the weight of $p$ between $0.1\times$ at 0.4~m/s and full weight at 1.0~m/s, i.e.\
\begin{equation}
  \small
  \bgscale{p} = \begin{cases}
    0.1      & \text{if } \pointspeed{p} < 0.4\text{ m/s} \\
    1.0      & \text{if } \pointspeed{p} > 1.0\text{ m/s} \\
    1.8s-0.8 & \text{o.w.}
  \end{cases}
  \equationlabel{pseudolabelscale}
\end{equation}
These thresholds are selected to down-weight approximately 80\% of points by $0.1\times$, with the other 20\% of points split between the soft  and full weight region\footnote{For Argoverse~2, exactly 78.1\% of points are downweighted, 11.8\% lie in the soft-weight region, and 10.1\% lie in the full weight region; for Waymo Open 80.0\% of points are downweighted, 7.9\% lie in the soft-weight region, and 12.1\% lie in the full-weight region respectively.}. In \tableref{argochodoshextended}, we show that our weighting scheme provides non-trivial improvements over uniform weighting (i.e.\ $\bgscale{\cdot} = 1$) for \ourmethodonex{}; however, it actually hurts performance for \ourmethodthreex{}.

\begin{table}[h]
  \centering
  \caption{Comparison between \ourmethod{} trained on Argoverse 2 using NSFP pseudo-labels and \ourmethod{} using \citet{chodosh2023} pseudo-labels using both uniform and speed scaled point weighting. Methods with an * have performance averaged over 3 training runs (see \appendixref{intertrainvariance} for details).}
  \scriptsize
  \setlength{\tabcolsep}{3pt}
  \begin{tabular}{lrrrr}
\toprule
& Threeway & Dynamic & Static & Static\\
& EPE & FG EPE & FG EPE & BG EPE \\
\midrule
\ourmethodonex{} (\equationref{pseudolabelscale}, NSFP pseudo-labels)* & $0.084$ & $0.217$ & $0.023$ & $0.011$\\
\ourmethodonex{} (\equationref{pseudolabelscale}, \citet{chodosh2023} pseudo-labels) & $0.086$ & $0.227$ & $0.019$ & $0.011$  \\
\ourmethodonex{}  (NSFP pseudo-labels)* 
& $0.088$ & $0.231$ & $0.022$ & $0.011$  \\
\ourmethodonex{} (\citet{chodosh2023} pseudo-labels)
& $0.085$ & $0.234$ & $0.018$ & $0.004$  \\ \midrule
\ourmethodxlthreex{} & $0.053$ & $0.131$ & $0.018$ & $0.011$ \\
\ourmethodxlthreex{} (\equationref{pseudolabelscale}) & $0.056$ & $0.139$ & $0.017$ & $0.011$ \\
\bottomrule
\end{tabular}
  \tablelabel{argochodoshextended}
\end{table}

\subsection{How much of FastFlow3D's performance is due to its semantic point weighting?}\sectionlabel{semanticsmatter}

Unlike \ourmethod{}, FastFlow3D \emph{can} use human foreground / background point labels to upweight the flow importance of foreground points (\sectionref{fastflow3d}, \equationref{bgscale}). To understand the impact of this weighting, we train FastFlow3D with two modified losses; rather than scaling using semantics as described in \equationref{bgscale}, we uniformly weight all points ($\bgscale{\cdot} = 1$) or our speed based weighting (\equationref{pseudolabelscale}).

\begin{table}[h]
  \centering
  \caption{Comparison between \ourmethod{}, FastFlow3D, and the ablated {FastFlow3D with uniform scaling ($\bgscale{\cdot} = 1$)} trained on Argoverse~2. The performance of FastFlow3D with Uniform Scaling and our speed scaling (\equationref{pseudolabelscale}) are nearly identical to \ourmethod{}'s performance.
  Methods with an * have performance averaged over 3 training runs (see \appendixref{intertrainvariance} for details).
  Underlined methods require human supervision.}
  \scriptsize
  \setlength{\tabcolsep}{3pt}
  \begin{tabular}{lrrrr}
    \toprule
& Threeway & Dynamic & Static  & Static  \\
& EPE      & FG EPE  & FG EPE  & BG EPE  \\ \midrule
    \ourmethodonex{}* (Ours)& $0.088$  & $0.231$ & $0.022$ & $0.011$   \\
    \humanlabels{FastFlow3D ($\bgscale{\cdot} = 1$)}               & $0.081$  & $0.220$ & $0.018$ & $0.006$ \\
    \humanlabels{FastFlow3D (\equationref{pseudolabelscale})} & $0.081$ & $0.224$ & $0.018$ & $0.002$ \\
    \humanlabels{FastFlow3D}*~\citep{scalablesceneflow} & $0.071$  & $0.186$ & $0.021$ & $0.006$ \\ 
    \bottomrule
  \end{tabular}
  \tablelabel{argoablation}
\end{table}

 As shown in \tableref{argoablation}, the performance of FastFlow3D ($\bgscale{\cdot} = 1$) and (\equationref{pseudolabelscale}) degrades more than halfway to \ourmethod{}'s performance.

This raises the question: why is the performance improvement of semantic weighting larger than the improvement of our unsupervised moving point weighting scheme (\appendixref{speedscaling})? We conjecture that not only does semantic weighting provide increased loss on moving objects, it implicitly teaches the network to recognize the structure of objects themselves. For example, with \equationref{bgscale} scaling, end-point error on a stationary pedestrian is significantly higher than static background points, incentivizing the network to learn to detect the point \emph{structure} common to pedestrians, even if immobile, to perfect the predictions on those points.

\section{Characterizing inter-training run final performance variance for \ourmethod{} and FastFlow3D}\appendixlabel{intertrainvariance}

On Argoverse 2, Threeway EPE difference between \ourmethod{} and the human supervised FastFlow3D is $1.6$cm (\tableref{argobigtable}); how much of this gap can be attributed to training variance between runs? To answer this question, we train \ourmethod{} and FastFlow3D from scratch 3 times each. \ourmethod{} is trained on the same Argoverse 2 NSFP pseudo-labels (\tableref{argoretrain}), resulting in a mean Threeway EPE of $0.088$m with error of $0.003$m ($0.3$cm) in either direction, and FastFlow3D is trained on the Argoverse 2 human labels (\tableref{argoretrainfastflow}), resulting in a mean Threeway EPE of $0.071$m with error under $0.003$m ($0.3$cm) in either direction.

To contextualize the scale of this variance, the underlying Velodyne VLP-32 sensors used to collect the Argoverse 2 are only certified to $\pm 3$ cm of error~\citep{lasersystemcharacterization} (an order of magnitude greater than the deviation from the mean train performance for \ourmethod{}), and this entirely neglects additional sources of noise introduced from other real world effects such as empirical ego motion compensation.



\begin{table}[h]
  \centering
  \caption{Performance of \ourmethod{} over 3 train runs on the same NSFP pseudo-labels.}
  \scriptsize
  \setlength{\tabcolsep}{3pt}
  \begin{tabular}{lrrrr}
    \toprule
& Threeway & Dynamic & Static  & Static     \\
& EPE      & FG EPE  & FG EPE  & BG EPE  \\ \midrule
    \ourmethodonex{} Run \#1 & $0.085$  & $0.224$ & $0.021$ & $0.011$   \\
    \ourmethodonex{} Run \#2 & $0.088$  & $0.231$ & $0.022$ & $0.010$\\
    \ourmethodonex{} Run \#3 & $0.091$  & $0.240$ & $0.023$ & $0.011$ \\
    \midrule
    \ourmethodonex{} Average & $0.088$  & $0.231$ & $0.022$ & $0.011$ \\
    \bottomrule
  \end{tabular}
  \tablelabel{argoretrain}
\end{table}

\begin{table}[h]
  \centering
  \caption{Performance of \ourmethod{} ablated with point scaling  (\equationref{pseudolabelscale}) over 3 train runs on the same NSFP pseudo-labels.}
  \scriptsize
  \setlength{\tabcolsep}{3pt}
  \begin{tabular}{lrrrr}
    \toprule
& Threeway & Dynamic & Static  & Static    \\
& EPE      & FG EPE  & FG EPE  & BG EPE   \\ \midrule
    \ourmethodonex{} (\equationref{pseudolabelscale}) Run \#1 & $0.083$  & $0.214$ & $0.023$ & $0.011$    \\
    \ourmethodonex{} (\equationref{pseudolabelscale}) Run \#2 & $0.083$  & $0.215$ & $0.024$ & $0.011$    \\
    \ourmethodonex{} (\equationref{pseudolabelscale}) Run \#3 & $0.085$  & $0.222$ & $0.022$ & $0.011$    \\
    \midrule
    \ourmethodonex{} (\equationref{pseudolabelscale}) Average & $0.084$  & $0.217$ & $0.023$ & $0.011$    \\
    \bottomrule
  \end{tabular}
  \tablelabel{argoretrainneweq}
\end{table}

\begin{table}[h]
  \centering
  \caption{Performance of FastFlow3D over 3 train runs on the Argoverse 2 human labels.}
  \scriptsize
  \setlength{\tabcolsep}{3pt}
  \begin{tabular}{lrrrr}
    \toprule
& Threeway & Dynamic & Static  & Static \\
& EPE      & FG EPE  & FG EPE  & BG EPE \\ \midrule
    FastFlow3D Run \#1 & $0.070$  & $0.181$ & $0.020$ & $0.006$   \\
    FastFlow3D Run \#2 & $0.071$  & $0.186$ & $0.021$ & $0.007$ \\
    FastFlow3D Run \#3 & $0.073$  & $0.191$ & $0.023$ & $0.006$ \\
    \midrule
    FastFlow3D Average & $0.071$  & $0.186$ & $0.021$ & $0.006$ \\
    \bottomrule
  \end{tabular}
  \tablelabel{argoretrainfastflow}
\end{table}

\section{Characterizing how \ourmethod{}'s performance evolves during training}\appendixlabel{epochs_error}

Threeway EPE breaks down performance into three categories: \emph{Foreground Dynamic}, \emph{Foreground Static}, and \emph{Background}. How does \ourmethod{}'s performance evolve during training?

To understand this, we plot \ourmethodonex{} and \ourmethodthreex{} in \figref{threeway_epe_breakdown}. Both methods converge to their final background performance almost immediately, and most of the improvements seen in the final Threeway EPE stem from improvements in Foreground Dynamic (\figref{fg_dynamic_performance}). The impact of additional data is also made clear early in training, as \ourmethodthreex{} has significantly lower Threeway EPE by epoch 15 than \ourmethodonex{}.

\begin{figure}[htbp!]

  \centering
  \begin{subfigure}[T]{0.48\textwidth}
    \centering
    \includegraphics{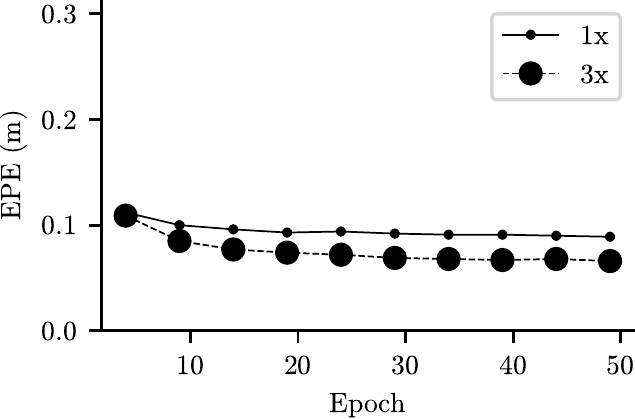}
    \caption{Threeway EPE}
    \figlabel{}
  \end{subfigure}
  \hfill
  \begin{subfigure}[T]{0.48\textwidth}
    \centering
    \includegraphics{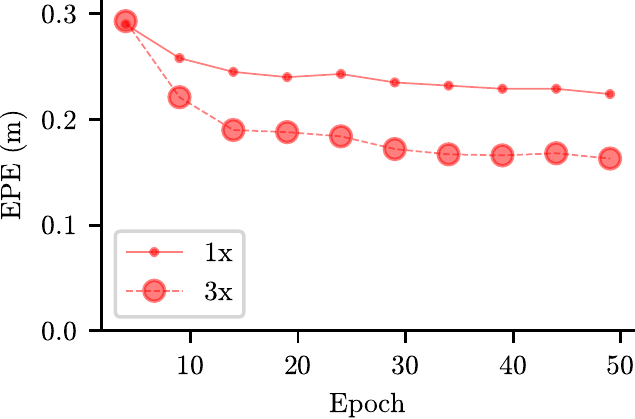}
    \caption{Foreground Dynamic}
    \figlabel{fg_dynamic_performance}
  \end{subfigure}

  \begin{subfigure}[T]{0.48\textwidth}
    \centering
    \includegraphics{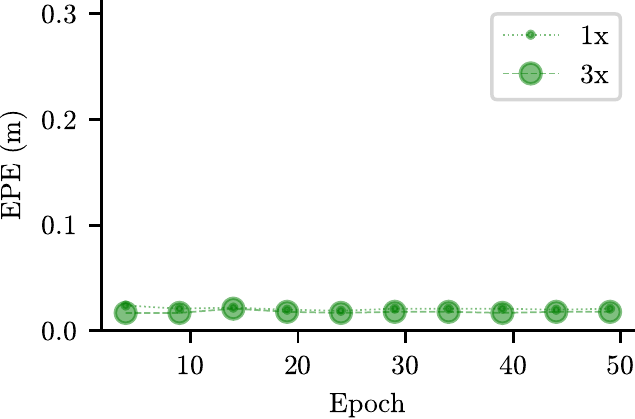}
    \caption{Foreground Static}
    \figlabel{}
  \end{subfigure}
  \hfill
  \begin{subfigure}[T]{0.48\textwidth}
    \centering
    \includegraphics{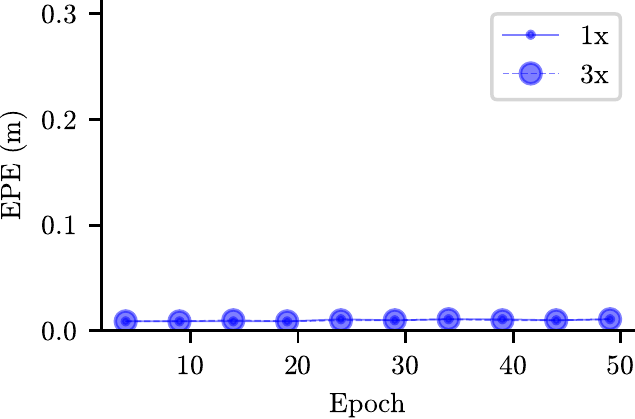}
    \caption{Background}
    \figlabel{}
  \end{subfigure}

  \caption{Performance of \ourmethodonex{} and \ourmethodthreex{} on the Argoverse~2 \emph{val} split by training epoch. Both methods converge to their final background performance almost immediately, and most of the improvements seen in the final Threeway EPE stem from improvements in Foreground Dynamic (\figref{fg_dynamic_performance}).}
  \figlabel{threeway_epe_breakdown}
\end{figure}

\section{Estimating Human Labeling versus Pseudo-labeling costs}\appendixlabel{labelvspseudolabelcosts}

NSFP pseudolabeling of the Argoverse 2 {train} split (700 sequences of 150 frames) required a total of 753 hours of NVidia Turing generation GPU time. At September, 2023 Amazon Web Services EC2 prices, a single \texttt{g4dn.xlarge}, equipped with a single NVidia Tesla T4, costs \$0.526 per hour\footnote{\url{https://aws.amazon.com/ec2/pricing/on-demand/}}, for a total cost of \$394 to pseudo-label. By comparison, at an estimated \$0.10 per frame per cuboid (no public cost statements exist for production quality AV dataset labels, but this the standard price point within the industry), Argoverse 2's train split has an average of 75 cuboids per frame~\citep{argoverse2}, for a total cost on the order of \$787,500 to human annotate.

\section{Details on Endpoint Residuals}\appendixlabel{endpoint_errors_details}

The process of constructing these endpoint residual plots is shown in \figref{endpoint_residual_construction}. For moving points (points with a ground truth flow vector magnitude >0.5m/s), the raw points (\figref{raw_points}) are transformed into a standard frame with the ground truth vector pointing up and the endpoint at the center of the plot (\figref{standard_frame}), and the residual endpoints are accumulated (\figref{error_dots}). Residual plots for baselines, as well as their unrotated counterparts, are shown in \figref{endpoint_distribution_baselines}.

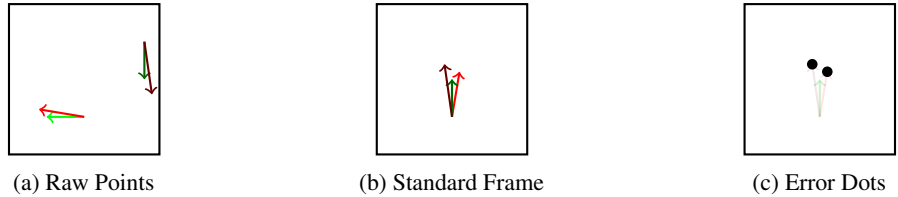
\begin{figure}
    \centering
    \begin{subfigure}[T]{0.30\textwidth}
    \centering
    \begin{tikzpicture}
    \definecolor{lightgreen}{rgb}{0,1,0}
    \definecolor{darkgreen}{rgb}{0,0.4,0}
    \definecolor{lightred}{rgb}{1,0,0}
    \definecolor{darkred}{rgb}{0.4,0,0}

    \draw[->, lightgreen, thick] (0.5,0) -- (0,0); 
    \draw[->, lightred, thick] (0.5,0) -- (-0.1,0.1);

    \draw[->, darkgreen, thick] (1.3,1) -- (1.3,0.5); 
    \draw[->, darkred, thick] (1.3,1) -- (1.4,0.3);
    
    \draw[thick] (-0.5,-0.5) rectangle (1.5,1.5);
    \end{tikzpicture}
    \caption{Raw Points}
    \figlabel{raw_points}
    \end{subfigure}
    \hfill
    \begin{subfigure}[T]{0.30\textwidth}
    \centering
    \begin{tikzpicture}
    \definecolor{lightgreen}{rgb}{0,1,0}
    \definecolor{darkgreen}{rgb}{0,0.4,0}
    \definecolor{lightred}{rgb}{1,0,0}
    \definecolor{darkred}{rgb}{0.4,0,0}
    \draw[->, lightgreen, thick] (0.5,0) -- (0.5,0.5);
    
    \draw[->, lightred, thick] (0.5,0) -- (0.6,0.6);

    \draw[->, darkgreen, thick] (0.5,0) -- (0.5,0.5);

    \draw[->, darkred, thick] (0.5,0) -- (0.4,0.7);
    
    \draw[thick] (-0.5,-0.5) rectangle (1.5,1.5);
    \end{tikzpicture}
    \caption{Standard Frame}
    \figlabel{standard_frame}
    \end{subfigure}
     \hfill
    \begin{subfigure}[T]{0.30\textwidth}
    \centering
    \begin{tikzpicture}
    \definecolor{lightgreen}{rgb}{0,1,0}
    \definecolor{darkgreen}{rgb}{0,0.4,0}
    \definecolor{lightred}{rgb}{1,0,0}
    \definecolor{darkred}{rgb}{0.4,0,0}
    \draw[->, lightgreen, thick, opacity=0.1] (0.5,0) -- (0.5,0.5);
    
    \draw[->, lightred, thick, opacity=0.1] (0.5,0) -- (0.6,0.6);

    \draw[->, darkgreen, thick, opacity=0.1] (0.5,0) -- (0.5,0.5);

    \draw[->, darkred, thick, opacity=0.1] (0.5,0) -- (0.4,0.7);

    \fill[black] (0.6,0.6) circle (2pt);

    \fill[black] (0.4,0.7) circle (2pt);
    
    \draw[thick] (-0.5,-0.5) rectangle (1.5,1.5);
    \end{tikzpicture}
    \caption{Error Dots}
    \figlabel{error_dots}
    \end{subfigure}
    
    \caption{Process for constructing the endpoint residual plots. The raw points (\figref{raw_points}) are transformed into a standard frame with the ground truth vector pointing up and the endpoint at the center of the plot (\figref{standard_frame}), and the residual endpoints are accumulated (\figref{error_dots}).}
    \figlabel{endpoint_residual_construction}
\end{figure}

\begin{figure}[htb!]

  \centering
  \begin{subfigure}[T]{0.22\textwidth}
    \centering
    \includegraphics{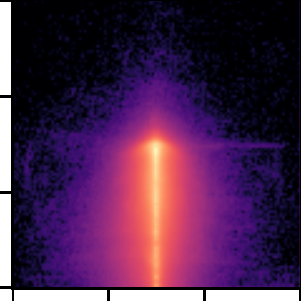}
    \caption{Nearest Neighbor, Log, Rotated}
    \figlabel{endpoint_distribution_nearest_neighbor}
  \end{subfigure}
  \hfill
  \begin{subfigure}[T]{0.22\textwidth}
    \centering
    \includegraphics{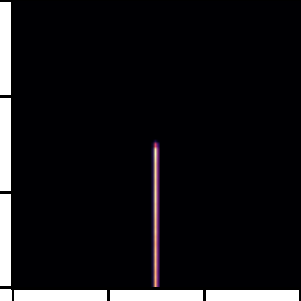}
    \caption{$\vec{0}$ Flow, Log, Rotated}
    \figlabel{endpoint_distribution_odom}
  \end{subfigure}
  \hfill
  \begin{subfigure}[T]{0.22\textwidth}
    \centering
    \includegraphics{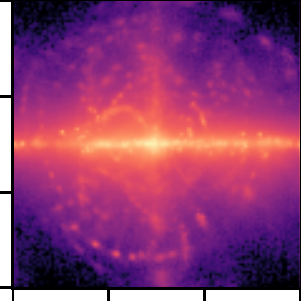}
    \caption{Nearest Neighbor, Log, Unrotated}
    \figlabel{endpoint_distribution_nearest_neighbor_unrotated}
  \end{subfigure}
  \hfill
  \begin{subfigure}[T]{0.22\textwidth}
    \centering
    \includegraphics{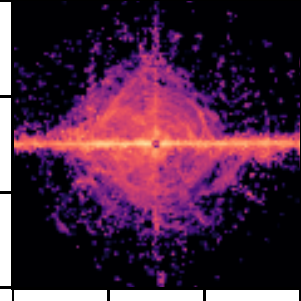}
    \caption{$\vec{0}$ Flow, Log, Unrotated}
    \figlabel{endpoint_distribution_odom_unrotated}
  \end{subfigure}

  \begin{subfigure}[T]{0.22\textwidth}
    \centering
    \includegraphics{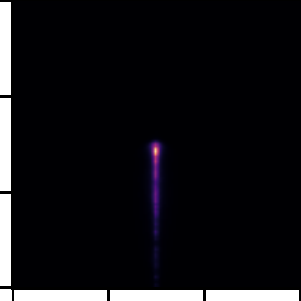}
    \caption{Nearest Neighbor, Abs, Rotated}
    \figlabel{endpoint_distribution_nsfp_unrotated}
  \end{subfigure}
  \hfill
  \begin{subfigure}[T]{0.22\textwidth}
    \centering
    \includegraphics{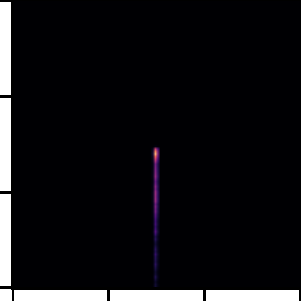}
    \caption{$\vec{0}$ Flow, Abs, Rotated}
    \figlabel{endpoint_distribution_nsfp_unrotated}
  \end{subfigure}
  \hfill
  \begin{subfigure}[T]{0.22\textwidth}
    \centering
    \includegraphics{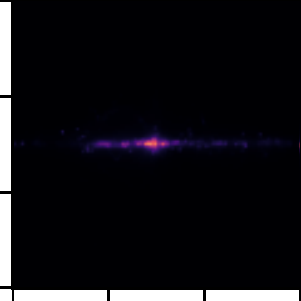}
    \caption{Nearest Neighbor, Abs, Unrotated}
    \figlabel{endpoint_distribution_nearest_neighbor_unrotated}
  \end{subfigure}
  \hfill
  \begin{subfigure}[T]{0.22\textwidth}
    \centering
    \includegraphics{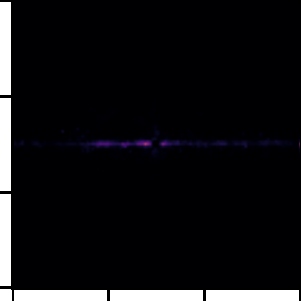}
    \caption{$\vec{0}$ Flow, Abs, Unrotated}
    \figlabel{endpoint_distribution_odom_unrotated}
  \end{subfigure}

  \caption{Birds-eye-view heatmap of endpoint residuals for na\"ive flow methods of predicting flow (Nearest Neighbor and $\vec{0}$ Flow on all points) for non-background points moving above 0.5m/s in the raw coordinate frame of the ground truth labels. Brighter color indicates more points in each bin. Perfect labels would produce a single central dot. Distance between ticks is 1 meter. Top row shows frequency on a log color scale to display error distribution shape. Bottom row shows frequency on an absolute color scale to display centroid. Left half shows results in the rotated ground truth coordinate frame. Right half shows results in the unrotated ground truth coordinate frame.}
  \figlabel{endpoint_distribution_baselines}
\end{figure}

\section{FAQ}

\subsection{Our method is ``just'' a combination of existing methods using standard distillation. Where does the novelty come in?}

Michael Black argues that ``the simplicity of an idea is often confused with a lack of novelty when exactly the opposite is often true.''~\citep{reviewerguidelines}.
Indeed, we think our novelty comes from the fact that our simple and post-hoc obvious pipeline produces surprisingly good results; our simple pipeline need only consume more raw data to improve and capture state-of-the-art over expensive human supervision while using the same feedforward model architectures. 

\subsection{What are the fundamental insights from this paper? What new knowledge was generated?}

Beyond producing a useful artifact, our straight-forward pipeline shows that simply training a supervised model with imperfect pseudo-labels can \emph{exceed} the performance of perfect human labels on substantial fraction of the data. We think this is itself surprising, but we also think it has highly impactful implications for the problem of scene flow estimation: \emph{point cloud quantity and diversity is more important than perfect flow label quality for training feedforward scene flow estimators}.

We also think this statement and our empirical scaling laws~(\sectionref{scalinglaws}) lead directly to actionable advice for practitioners at Autonomous Vehicle companies and other organizations with a large trove of diverse point cloud data: \emph{scaling \ourmethod{} on this large scale data will net a significantly better scene flow estimator than expensive human supervision would using a 1000$\times$ larger budget}.

In addition to insights, we also present a novel scene flow estimation analysis technique. To our knowledge, the residual plots in \sectionref{noise_of_our_method} are the first attempt at visualizing the residual \emph{distribution} of scene flow estimators. We think these plots provide useful insights to practitioners and researchers, particularly for consumption in downstream tasks; as an example, open world object extraction~\citep{objectdetectionmotion} requires the ability to threshold for motion and cluster motion vectors together to extract the entire object. Decreased average EPE is useful for this task, but understanding the \emph{distribution} of flow vectors is needed to craft good extraction heuristics.

\end{document}